\documentclass[11pt,a4paper]{article}
\usepackage[hyperref]{emnlp-ijcnlp-2019}
\usepackage{times}
\usepackage{latexsym}
\usepackage{graphicx}
\usepackage{lipsum}
\usepackage[utf8]{inputenc}
\usepackage{amsmath}
\usepackage{array}
\usepackage{url}
\usepackage{float}
\usepackage{xcolor}

\usepackage{booktabs}
\usepackage{url}

\aclfinalcopy 

\title{SmokEng: Towards Fine-grained Classification of Tobacco-related Social Media Text}

\author{Kartikey Pant, Venkata Himakar Yanamandra, Alok Debnath and Radhika Mamidi \\
  International Institute of Information Technology \\
Hyderabad, Telangana, India \\
  \{kartikey.pant, himakar.y, alok.debnath\}@research.iiit.ac.in \\
  radhika.mamidi@iiit.ac.in
  }

\date{}

\begin{document}
\maketitle
\begin{abstract}
Contemporary datasets on tobacco consumption focus on one of two topics, either public health mentions and disease surveillance, or sentiment analysis on topical tobacco products and services. However, two primary considerations are not accounted for, the language of the demographic affected and a combination of the topics mentioned above in a fine-grained classification mechanism. In this paper, we create a dataset of 3144 tweets, which are selected based on the presence of colloquial slang related to smoking and analyze it based on the semantics of the tweet. Each class is created and annotated based on the content of the tweets such that further hierarchical methods can be easily applied.
    
Further, we prove the efficacy of standard text classification methods on this dataset, by designing experiments which do both binary as well as multi-class classification. Our experiments tackle the identification of either a specific topic (such as tobacco product promotion), a general mention (cigarettes and related products) or a more fine-grained classification. This methodology paves the way for further analysis, such as understanding sentiment or style, which makes this dataset a vital contribution to both disease surveillance and tobacco use research.

\end{abstract}

\section{Introduction}

\begin{figure*}[]
\centering
\includegraphics[width=16cm]{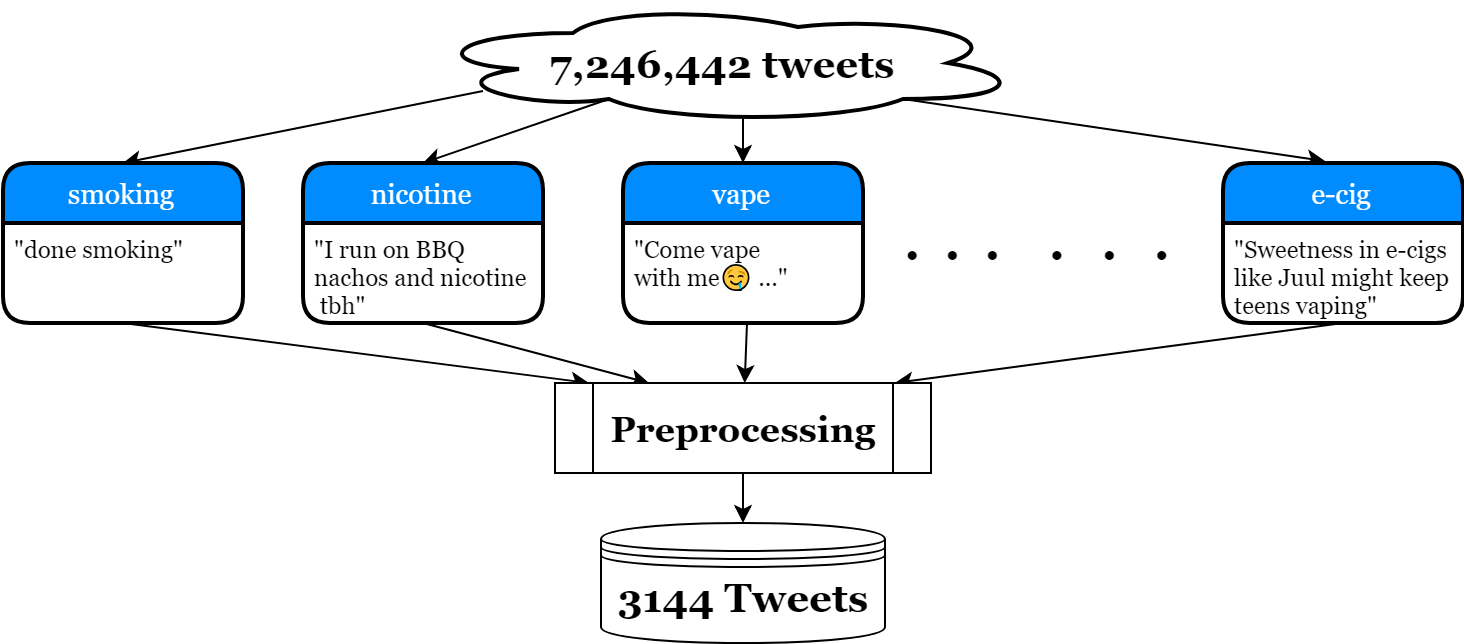}
\caption{Procedure for Data Collection. We started out with approximately 7 million Tweets which were mined based on 24 slang terms. These were pre-processed to select relevant tweets with decent traction on Twitter. A final cleaned dataset of 3144 tweets is presented.}
\end{figure*}
As Twitter has grown in popularity to 330 million monthly active users, researchers have increasingly been using it as a source of data for tobacco surveillance \citep{lienemann2017}. Tobacco-related advertisements, tweets, awareness posts, and related information is most actively viewed by young adults (aged 18 to 29), who are extensive users of social media and also represent the largest population of smokers in the US and Canada \footnote{\url{https://www.cdc.gov/tobacco/data_statistics/fact_sheets/adult_data/cig_smoking/}}. Furthermore, it allows us to understand patterns in ethnically diverse and vulnerable audiences \citep{lienemann2017}. Social media provides an active and useful platform for spreading awareness, especially dialog platforms, which have untapped potential for disease surveillance \citep{platt2016}. These platforms are useful in stimulating the discussion on societal roles in the domain of public health \citep{platt2016}. \citet{sharpe2016} has shown the utility of social media by highlighting that the number of people using social media channels for information about their illnesses before seeking medical care.

Correlation studies have shown that the most probable leading cause of preventable death globally is the consumption of tobacco and tobacco products \citep{prochaska2012}. The disease most commonly associated with tobacco consumption is lung cancer, with two million cases reported in 2018 alone \footnote{\url{https://www.wcrf.org/dietandcancer/cancer-trends/lung-cancer-statistics}}. While cigarettes are condemned on social media, this has been rivaled by the rising popularity and analysis of the supposed benefits of e-cigarettes \citep{dai2017mining}. Information pertaining to new flavors and innovations in the industry and surrounding culture have generated sizable traffic on social media as well \citep{Hiltone2016}. Studies show that social acceptance is a leading factor to the use and proliferation of e-cigarettes, with some reports claiming as many as 2.39 million high school and 0.63 million middle school students having used an e-cigarette at least once \citep{malik2019, mantey2019retail}. However, there are strong claims suggesting the use of e-cigarettes as a 'gateway' drug for other illicit substances \citep{Unger2016}. 


In this paper, we aim at classifying tweets relating to cigarettes, e-cigarettes, and other tobacco-related products into distinct classes. This classification is fine-grained in order to assist in the analysis of the type of tweets which affect the users the most for each product or category. The extensive, manually annotated dataset of 3144 tweets pertains to tobacco use classification into advertisement, general information, personal information, and non-tobacco drug classes. Such classification provides insight into the type of tweet and associated target audience. For example, present cessation programs target users who are ready to quit rather than people who use it regularly, which can be solved using twitter and other online social media \citep{prochaska2012}. Unlike many previous studies, we also include common slang terms into the classification scheme so as to be able to work with the social media discourse of the target audience.

Finally, we present several text-classification models for the fine-grained classification tasks pertaining to tobacco-related tweets on the released dataset\footnote{\url{https://github.com/kartikeypant/smokeng-tobacco-classification}}. In doing so, we extend the work in topical Twitter content analysis as well as the study of public health mentions on Twitter.


\begin{table*}[]
\centering
\begin{tabular}{|c|c|c|}
\hline
\textbf{Name}                            & \textbf{Label} & \textbf{\begin{tabular}[c]{@{}c@{}}Annotation \\ Class\end{tabular}} \\ \hline
\textit{Mention of Non-Tobacco Drugs}    & OD             & -1                                                                    \\ \hline
\textit{Unrelated or Ambiguous Mention} & UM             & 0                                                                    \\ \hline
\textit{Personal or Anecdotal Mention}   & PM             & 1                                                                    \\ \hline
\textit{Informative or Advisory Mention} & IM             & 2                                                                    \\ \hline
\textit{Advertisements}                  & AD             & 3                                                                    \\ \hline
\end{tabular}
\caption{Label and ID associated with each class.}
\label{tab: annotationClassMap}
\end{table*}

\section{Related Work}

\citet{myslin2013} explored content and sentiment analysis of Tobacco-related Twitter posts and performed analysis using machine learning classifiers for the detection of tobacco-relevant posts with a particular focus on emerging products like e-cigarettes and hookah. Their work depends on a triaxial classification along and uses basic statistical classifiers. However, their feature-engineered keyword-based systems do not account for slang associated with tobacco consumption. 


\citet{vandewater2018whose} performs a classification study based on identifying brand associated with a post using basic text analytics using keywords and image-based classifiers to determine the brands that were most responsible to posting about their brands on social media. \citet{cortese2018smoking} does a similar analysis on the consumer side, for female smokers on Instagram, targeting the same age group, but based entirely on feature extraction on images, particularly selfies. 

More recently, \citet{malik2019} explored patterns of communication of e-cigarette company Juul use on Twitter. They categorized 1008 randomly selected tweets across four dimensions, namely, user type, sentiment, genre, theme. However, they explore the effects of only Juul, and not other cigarettes or e-cigarettes, further limiting their experiment to only Juul-based analysis and inferences.

In the domain of Disease Surveillance, \citet{aramaki2011twitter} explored the problem of identifying influenza epidemics using machine-learning based tweet classifiers along with search engine trends for medical keywords and medical records for the disease in a local environment. For doing so, they use SVM based classifiers for extracting tweets that mention actual influenza patients. However, since they use only SVM based classifiers, they are limited in their accuracy in classification.

\citet{dai2017social} also focuses on public health surveillance, and uses word embeddings on a topic classifier in order to identify and capture semantic similarities between medical tweets by disease and tweet type for a more robust yet very filtered classification, not accounting for the variety of linguistic features in tweets such as slang, abbreviations and the like in the keyword-based classification mechanism. \citet{jiang2018identifying} works on a similar problem using machine learning solutions such as an LSTM classifier.

\section{Dataset Creation}

In this section, we explain the development of the dataset that we present along with this paper. We summarize the methods for collecting and filtering through the tweets to arrive at the final dataset and provide some examples of the types of tweets and features we focused on. We also provide the dataset annotation schema and guidelines.

\subsection{Data Collection}
Using the Twitter Application Programming Interface (API\footnote{https://developer.twitter.com/en/products/tweets/sample.html}), we collected a sample of tweets between 1st October 2018 and 7th October 2018 that represented 1\% of the entire Twitter feed. This 1\% sample consisted of an average 1,035,206 million tweets per day. Out of the 7,246,442 tweets, only tweets written in English and written by users with more than 100 followers have selected for the next step in order to clear spam written by bots.

In order to extract tobacco related tweets from this dataset, we constructed a list of keywords relevant to general tobacco usage, including hookah and e-cigarettes. Our initial list consisted of 32 such terms compiled from online slang dictionaries, but we pruned this list to 24 terms.
These were \textit{smoking, cigarette, e-cig*, cigar, tobacco, hookah, shisha, e-juice, e-liquid, vape, vaping, cheroot, cigarillo, roll-up, ashtray, baccy, rollies, claro, chain-smok*, vaper, ciggie, nicotine, non-smoker, non-smoking}.

By taking the dataset for a full week, we thus avoided potential bias based on the day of the week, which has been observed for alcohol related tweets, which spike in positive sentiment on Fridays and Saturdays \citep{CavazosRehg2015}. For each of the 7 days, all tweets matching any of the listed keywords were included. Tweets matching these tobacco related keywords reflected 0.00043\% of all tweets in the Twitter API 1\% sample. The resulting final dataset thus contained 3144 tweets, with a mean of 449 tweets per day.

\begin{table*}[]
\begin{tabular}{|c|l|}
\hline
\multicolumn{1}{|l|}{\textbf{Label}} & \multicolumn{1}{c|}{\textbf{Examples}}                                                                                                                                                                                    \\ \hline
UM                                      & \textit{\begin{tabular}[c]{@{}l@{}}"What are you smoking bruh ?"\\ "The smoking gun on Kavanaugh! $URL$ "\end{tabular}}                                                                                                   \\ \hline
PM                                      & \textit{\begin{tabular}[c]{@{}l@{}}"im smoking and doing whats best for me"\\ "I haven’t had a cigarette in \$NUMBER\$ months why do I want one so bad now??"\end{tabular}}                                               \\ \hline
IM                                      & \textit{\begin{tabular}[c]{@{}l@{}}"Obama puffed. Clinton did cigar feel.Churchill won major wars on whisky."\\ "The FDA’s claim of a teen vaping addiction epidemic doesn’t add up. \#ecigarette \#health"\end{tabular}} \\ \hline
AD                                      & \textit{\begin{tabular}[c]{@{}l@{}}"Which ACID Kuba Kuba are you aiming for? \#De4L \#ExperienceAcid \#cigar \#cigars $URL$" \\ "Spookah Lounge: A concept - a year round Halloween-themed hookah lounge"\end{tabular}}   \\ \hline
OD                                      & \textit{\begin{tabular}[c]{@{}l@{}}"Making my money and smoking my weed"\\ "Mobbin in da Bentley smoking moonrocks."\end{tabular}}                                                                                        \\ \hline
\end{tabular}
\caption{Examples for each category represented by its label.}
\end{table*}  

\subsection{Data Annotation}

The collected data was then annotated based on the categories mentioned in \autoref{tab: annotationClassMap}. These categories were chosen on the basis of frequency of occurrence, motivated by the general perception of tobacco and non-tobacco drug related tweets. These included advertisements as well anecdotes, information and cautionary tweets. We further noticed that a similar pattern was seen for e-cigarettes and also pertained to some other drugs. While we have explored e-cigarettes in this classification, we have marked the mention of other drugs that were tagged with the same keywords. 

A formal definition of each of the categories is given below.  

\begin{itemize}
    \item \textbf{Unrelated or Ambiguous Mention}: This category of tweets contain tweets containing information unrelated to tobacco or any other drug, or pertaining to ambiguity in the intent of the tweet, such as sarcasm. 
    
    \item \textbf{Personal or Anecdotal Mention}: Tweets are classified as containing a personal or anecdotal mention if they imply either personal use of tobacco products or e-cigarettes, or provide instances of use of the products by themselves or others.
    \item \textbf{Informative or Advisory Mention}: This class of tweets consist of a broad range of topics such as:
        \begin{itemize}
            \item mention or discussion on statistics of tobacco and e-cigarette use or consumption
            \item mention associated health risks or benefits
            \item portray the use of tobacco products or e-cigarettes by a public figure
            \item emphasize social campaigns for anti-smoking, smoking cessation and related products such as patches
        \end{itemize}
    \item \textbf{Advertisements}: All tweets written with the intent of the sale of tobacco products, e-cigarettes and associated  products or services are marked advertisements. In this classification, intent is considered using the mention of price as an objective measure.
    \item \textbf{Mention of Non-Tobacco Drugs}: Tweets which mention the use, sale, anecdotes and information about drugs other than e-cigarettes or tobacco products are annotated in this category.
\end{itemize}


\subsection{Inter-annotator Agreement}
Annotation of the dataset to detect the presence of tobacco substance use was carried out by two human annotators having linguistic background and proficiency in English. A sample annotation set consisting of 10 tweets per class was selected randomly from all across the corpus. Both annotators were given the selected sample annotation set. These sample annotation set served as a reference baseline of each category of the text. 

In order to validate the quality of annotation, we calculated the Inter-Annotator Agreement (IAA) for the fine-grain classification between the two annotation sets of 3,144 tobacco-related tweets using Cohen’s Kappa coefficient \citep{fleiss1973}. The Kappa score of \textbf{0.791} indicates that the quality of the annotation and presented schema is productive.

\section{Methodology}

In this section we describe the classifiers designed for this task of fine grained classification. The classifier architecture is based upon a combination of choosing word representations, along with a discriminator that is compatible with that representation. We use the TF-IDF for the suport vector machines and GloVe embeddings \citep{pennington2014glove} with our convolutional neural network architecture and recurrent architectures (LSTM and Bi-LSTM). We also used FastText and BERT embeddings (both base and large) with their native classifiers to note the change in the accuracies.

\subsection{Support Vector Machines (SVM)}

The first learning model used for classification in our experiment was Support Vector Machines (SVM) \citep{Cortes1995}. We used term frequency-inverse document frequency (TF-IDF) as a feature to classify the annotated tweets in our data set \citep{Salton1988}. TF-IDF captures the importance of the given the word in a document, defined in \autoref{eqn:tfidf}.

\begin{equation}
t f i d f(t, d, D)=f(t, d) \times \log \frac{N}{|\{d \epsilon D : t \epsilon d\}|}
\label{eqn:tfidf}
\end{equation}
where $f(t,d)$ indicates the number of times term $t$ appears in context, $d$ and $N$ is the total number of documents $|d \epsilon D : t \epsilon d|$ represents the total number of documents where $t$ occurs.

The SVM classifier finds the decision boundary that maximizes the margin by minimizing $||\textbf{w}||$ to find the optimal hyperplane for all the classification tasks:
\begin{equation}
\begin{array}{c}{\min f : \frac{1}{2}\|\mathbf{w}\|^{2}} \\ \\
{\text { s.t. } \quad y^{(i)}\left(\mathbf{w}^{T} \mathbf{x}^{(i)}+b\right) \geq 1, \quad i=1, \ldots, m}
\end{array}    
\end{equation}

where $\textbf{w}$ is the weight vector, $\textbf{x}$ is the input vector and $b$ is the bias.

\subsection{Convolutional Neural Networks (CNN)}

In this subsection, we outline the Convolutional Neural Networks \citep{Fukushima1988} for classification and also provide the process description for text classification in particular. Convolutional neural networks are multistage trainable neural networks architectures developed for classification tasks \citep{Lecun1998}. Each of these stages consist of the types of layers described below:
\begin{itemize}
    \item \textbf{Embedding Layer}: The purpose of an embedding layer is to transform the text inputs into a form which can be used by the CNN model. Here, each word of a text document is transformed into a dense vector of fixed size.
    
    \item \textbf{Convolutional Layers}: A Convolutional layer consists of multiple kernel matrices that perform the convolution mathematical operation on their input and produce an output matrix of features upon the addition of a bias value.
    
    \item \textbf{Pooling Layers}: The purpose of a pooling layer is to perform dimensionality reduction of the input feature vectors. Pooling layers use sub-sampling to the output of the convolutional layer matrices combing neighbouring elements. We have used the commonly used max-pooling function for the pooling.
    
    \item \textbf{Fully-Connected Layer}: It is a classic fully connected neural network layer. It is connected to the Pooling layers via a Dropout layer in order to prevent overfitting. Softmax activation function is used for defining the final output of this layer.
\end{itemize}
The following objective function is commonly used in the task:
\begin{equation}
    E_{w} = \frac{1}{n}\sum_{p=1}^{P}\sum_{j=1}^{N_{L}}(o_{j,p}^{L} - y_{j,p})^{2}
\end{equation} 
where $P$ is the number of patterns, $o_{j,p}^{L}$ is the output of $j^{th}$ neuron that belongs to $L^{th}$ layer, $N_{L}$ is the number of neurons in output of $L^{th}$ layer, $y_{j,p}$ is the desirable target of $j^{th}$ neuron of pattern $p$ and $y_i$ is the output associated with an input vector $x_i$ to the CNN. 

We use Adam Optimizer \citep{Kingma2014} to minimize the cost function $E_{w}$. 

\begin{table*}
\centering
\begin{tabular}{ccc}
\hline
\textbf{Model/Experiment} & \textbf{Personal Health Mentions} & \textbf{Tobacco-related Mentions} \\ \hline
~ $SVM$                    & 82.17\%                           & 83.44\%                         \\
~ $CNN$                    & 84.08\%                           & 82.48\%                         \\
~ $LSTM$                   & 84.39\%                           & 83.32\%                         \\
~ $BiLSTM$                 & 83.92\%                           & 82.97\%                         \\
~ $FastText$               & 83.76\%                           & 81.05\%                         \\ 
~ $BERT_{Base}$               & 85.19\%                           & 85.50\%                         \\
~ $BERT_{Large}$               & \textbf{87.26\%}                           & \textbf{85.67}\%                         \\ \hline
\end{tabular}
\caption{Binary Classification accuracies for specific topic (Personal Health Mention) or general theme (Tobacco-related Mentions).}
\label{tab:phm-ato}
\end{table*}

\subsection{Recurrent Neural Architectures}
Recurrent neural networks (RNN) have been employed to produce promising results on a variety of tasks, including language model and speech recognition \citep{Mikolov2010, Mikolov2011, Graves2005}. An RNN predicts the current output conditioned on long-distance features by maintaining a memory based on history information. 

An input layer represents features at time $t$. One-hot vectors for words, dense vector features such as word embeddings, or sparse features usually represent an input layer. An input layer has the same dimensionality as feature size. An output layer represents a probability distribution over labels at time $t$ and has the same dimensionality as the size of the labels. Compared to the feed-forward network, an RNN contains a connection between the previous hidden state and current hidden state. This connection is made through the recurrent layer, which is designed to store history information. The following equation is used to compute the values in the hidden, and output layers:
\begin{equation}
    \textbf{h}(t) = f(\textbf{Ux}(t) + \textbf{Wh}(t - 1)).
\end{equation}
\begin{equation}
    \textbf{y}(t) = g(\textbf{Vh}(t)),
\end{equation}

where $U$, $W$, and $V$ are the connection weights to be computed during training, and $f(z)$ and $g(z)$ are sigmoid and softmax activation functions as follows.
    \begin{equation}
    f(z) = \frac{1}{1 + e^{-z}},
\end{equation}
\begin{equation}
    g(z_{m}) = \frac{e^{z_{m}}}{\sum_{k}e^z_{k}}
\end{equation}
In this paper, we apply Long Short Term Memory (LSTM) and Bidirectional Long Short Term Memory(Bi-LSTM) to sequence tagging \citep{Hochreiter1997, Graves2005,Graves2013}. 

LSTM networks use purpose-built memory cells to update the hidden layer values. As a result, they may be better at finding and exploiting long-range dependencies in the data than a standard RNN. The following equation implements the LSTM model:
\begin{equation}
    i_{t} = \sigma(W_{xi}x_{t} + W_{hi}h_{t-1} + W_{ci}c_{t-1} + b_{i})
\end{equation}
\begin{equation}
    f_{t} = \sigma(W_{xf}x_{t} + W_{hf}h_{t-1} + W_{cf}c_{t-1} + b_{f} ) 
\end{equation}
\begin{equation}
    o_{t} = \sigma(W_{xo}x_{t} + W_{ho}h_{t-1} + W_{co}c_{t} + b_{o}) 
\end{equation}
\begin{equation}
    h_{t} = o_{t}tanh(c_{t})
\end{equation}

In sequence tagging task, we have access to both past and future input features for a given time. Thus, we can utilize a bidirectional LSTM network (Bi-LSTM) as proposed in \citep{Graves2013}. 

\begin{table*}
\centering
\begin{tabular}{llll} 
\hline
\textbf{~ Methods~~} & \textbf{~ Accuracy~~} & ~ \textbf{F1 Score}~~ & ~ \textbf{Recall~}~  \\ 
\hline
~ $SVM$       & ~ 65.45\%             & ~ 0.678               & ~ 0.657              \\
~ $CNN$                & ~ 66.72\%             & ~ 0.668               & ~ 0.599              \\
~ $LSTM$      & ~ 64.97\%             & ~ 0.641               & ~ 0.583              \\
~ $BiLSTM$    & ~ 65.29\%             & ~ 0.643               & ~ 0.597              \\
~ $FastText$  & ~ 69.43\%            & ~ 0.696               & ~ 0.669            \\
~ $BERT_{Base}$  & ~ 70.86\%            & ~ 0.708               & ~ 0.709              \\
~ $BERT_{Large}$  & ~ \textbf{71.34\%}             & ~ \textbf{0.714}               & ~ \textbf{0.713}              \\

\hline
\end{tabular}
\caption{Evaluation scores for the Fine-grained classification experiment.}
\label{tab:fin}
\end{table*}

\subsection{FastText}

FastText classifier has proven to be efficient for text classification  \citep{Joulin2016}.   It  is  often  at  par with deep learning classifiers  in  terms  of  accuracy, and much faster for training and evaluation. FastText  uses  bag  of  words  and  bag  of  n-grams as features for text classification. Bag of n-grams feature captures partial information about the lo-cal  word  order.  FastText  allows updating word vectors through back-propagation during training allowing the model to fine-tune word representations  according  to  the  task  at  hand \citep{Bojanowski2016}. The model is trained using stochastic gradient descent and a linearly decaying learning rate.

\subsection{BERT}
While previous studies on word representations focused on learning context-independent representations, recent works have focused on learning contextualized word representations. One of the more recent contextualized word representation is BERT \citep{Devlin2019}. 

BERT is a contextualized word representation model, pre-trained using bidirectional transformers\citep{Vaswani2017}. It uses a masked language model that predicts randomly masked in a sequence. It uses the task of \textit{next sentence prediction} for learning the embeddings with a broader context. It outperforms many existing techniques on most NLP tasks with minimal task-specific architectural changes. It is pretrained on 3.3B words from various sources including BooksCorpus and the English Wikipedia.

Based on the transformer architecture used, BERT is classified into two types: $BERT_{Base}$ and $BERT_{Large}$. $BERT_{Base}$ uses a 12-layered transformer with 110M parameters. $BERT_{Large}$ uses a 24-layered transformer with 340M parameters. We use the cased variant of both models.

\section{Experiments}


In this section, we describe three experiments on the dataset created in the section above. The experiments are designed to show how well existing models perform on the naive binary classification based on this dataset as well as the fine-grained five-class classification system. The first experiment is based on detecting just personal or anecdotal mentions. The second is based on identifying whether a tweet is about tobacco or not. The last experiment is a full fine-grained classification experiment.

The following experiments were conducted keeping an 80-20 split between training and test data, with 2517 tweets in the training dataset and 629 tweets in the test dataset. All tweets were shuffled randomly before the train-test split.

$BERT_{Large}$ was observed to perform the best in all three experiments, followed closely by $BERT_{Base}$ in all the experiments that were conducted.

\subsection{Experiment 1: Detecting Personal Mentions of Tobacco Use}

The first experiment in the study was to detect tweets containing personal mentions of tobacco use. Tweets containing personal mentions of tobacco use are the ones marking implicit or explicit use of a tobacco substance by the poster. The objective of this experiment is to analyze the best method to identify tweets which talk about tobacco in an anecdotal manner, which can be used to understand the semantic similarity between such tweets. \autoref{tab:phm-ato} illustrates the results for this experiment.

\subsection{Experiment 2: Identifying Tobacco-related Mentions}

The next experiment in the study was to detect all tobacco-related tweets related. These include the following categories of tweets: personal mentions of tobacco-use, general information about tobacco or its use, advertisements. Thus, the experiment was to determine whether the tweet belonged to one of the above categories or not. The objective here is also to gauge semantic information in tweets with mentions of tobacco, suggesting that tweets using the similar slang might be talking about other drugs or ambiguous or unrelated information. \autoref{tab:phm-ato} illustrates the results for this experiment.


\subsection{Experiment 3: Performing Fine-grained Classification of Tobacco-related Mentions}

The last experiment conducted in the study was to classify the tweets into all five categories: UM, PM, IM, AD, OD. Table \ref{tab:fin} illustrates the results of the experiment. This is essentially the fine grained classification experiment which relies on semantic information as well as lexical choice. We see that models from all the three experiments perform differently given the type of task. \autoref{tab:fin} illustrates the results for this experiment.





\section{Discussion}
In this section, we analyze our contributions from the perspective of advancing work in the fields of topical content analysis as well as the study of public health mentions in tweets, with regards to tobacco products, as well as e-cigarettes and related products. Given the effects of both as well as the significant overlap in the demographic of consumers of tobacco products and Twitter users, we found it necessary to understand the nature of the tweets produced and consumed by them.

\begin{figure}
    \centering
    \includegraphics[width=0.9\columnwidth]{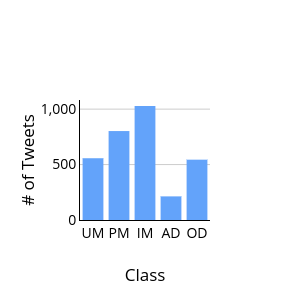}
    \caption{Distribution of tweets among different categories}
    \label{fig:class-distribution}
\end{figure}

Our dataset, a collection of 3144 tweets, accumulated and filtered over the period of just a week, implies that tobacco and related drugs are tweeted about and spoken of quite frequently, but the linguistic cues common among these tweets was not considered until now. The inclusion of tweets into the corpus based on slang terminology is an attempt to analyze the Twitter landscape in the language of the audience which most highly correlates with the demographic of consumers for the aforementioned products. To the best of our knowledge, using common slang as a basis of dataset creation and filtration for this task has not been attempted before.

\begin{table}[t]
\centering
\begin{tabular}{lll} 
\toprule
\textbf{Category} & \textbf{Retweets} & \textbf{Favorites}  \\ 
\hline
\textit{UM}    & 1079.05           & 0.794               \\
\textit{PM}    & 12171.60          & 0.904               \\
\textit{IM}    & 680.24            & 3.918               \\
\textit{AD}    & 140.81            & 4.586               \\
\textit{OD}    & 873.08            & 0.868               \\
\bottomrule
\end{tabular}
\caption{Average retweets and favorites across classes}
\label{tab:Stats}
\end{table}

Contemporary methods in the field focus on two basic characterizations, user based and sentiment based. User based classification such as \citet{malik2019} and \citet{jo2016price} are based on the analyzing activity from a particular user or set of users, while sentiment based analyses such as \citet{paul2017, allem2018} and \citet{myslin2013} are based on understanding the sentiment of the users on the basis of a new product, category or a more generalized perception of smoking in general. On the other hand, public health mention research such as \citet{jawad2015social} focuses on effect of a particular type of tweet, generally health campaigns. Fundamentally, the classes we have chosen for the collected data are based on the same principle as the data collection mechanism, with the aim to bridge the gap between the classification studies and the public health surveillance research. This is because our categories cover the breadth of the tweets evenly, directed towards semantically understanding the nature of the tweets. This information is vital for addressing the validity and reach of campaigns, advertisements and other efforts.




\autoref{fig:class-distribution} shows the distribution of the number of tweets in each class. We see that in the span of a week, informative or advisory and personal mentions are the most widely posted. The tweets that provide general information about smokers or the habits of smoking tobacco or e-cigarettes are generated the most, implying that a larger section of the population tweets of smoking in an anecdotal manner. Similarly, \autoref{tab:Stats} shows an interesting trends for the favorites. Advertisements have a higher average favorite count than most other classes, while anecdotal and advisory tweets are the most retweeted on average. This difference is an interesting observation, primarily because on further work such as sentiment analysis and doing short text style transfer \citep{luo2019towards} for these categories may provide an effective strategy for advertisers and campaigners alike.


\section{Conclusion and Future Work}

In this paper, we created a dataset of tweets and classified them in order to understand the social media atmosphere around tobacco, e-cigarettes and other related products. Our schema for categorization targets posts on public health as much as tobacco related products, therefore allowing us to know the number and type of tweets used in public health surveillance for the above mentioned products. Most importantly, we consider slang as a very important aspect of our data collection mechanism, which has allowed us to factor in the content which is circulated and exposed to the majority of the consumers of social media and the aforementioned products both. 

This contribution can be further extended by working with other social media platforms, where the methods introduced above can be easily replicated. Social media specific slang can be taken into account to make a more robust dataset for this task. Furthermore, on the public health surveillance aspect, more metadata using the tweets can be extracted, which gives an idea of the type of tweets or posts needed to grab the attention of a wider audience on topics of public health and awareness for the grave topic of tobacco products and e-cigarettes.  





\bibliography{emnlp-ijcnlp-2019}
\bibliographystyle{acl_natbib}

\appendix

\end{document}